%% file: main.tex
\documentclass{MITcsail}

\input{tex/command/math_commands}

\input{tex/command/mlVecMat}

\input{tex/command/box}

\usepackage{xcolor}
\definecolor{brandpink}{HTML}{FF2882} 

\usepackage{hyperref}
\usepackage{tikz}
\usepackage{xcolor}
\hypersetup{
    colorlinks=true,
    linkcolor=brandpink,
    citecolor=brandpink,
    filecolor=darkred,
    urlcolor=brandpink,
}
\usepackage[numbers, sort&compress]{natbib}
\usepackage{supertabular}

\usepackage{booktabs}
\usepackage{siunitx}
\usepackage{threeparttable}
\title{When Hallucination Costs Millions: \\[0.3em] Benchmarking AI Agents in High-Stakes Adversarial Financial Markets}

\author
{Zeshi Dai~$^{1,*}$, Zimo Peng~$^{1,*}$, Zerui Cheng~$^{2,*}$, Ryan Yihe Li~$^{1,\dagger}$\\
\vspace{1em}
\normalfont{\small $^{1}$ Surf AI, Cybertino Lab} \quad\quad
\normalfont{\small $^{2}$ Princeton University}\\
\vspace{1em}
\texttt{Link: 
\href{https://huggingface.co/spaces/cyberco/CAIA_0927_Leaderboard}{Leaderboard}
~|~
\href{https://github.com/caiba-ai/caia-benchmark-0927}{Github Repository}
~|~
\href{https://huggingface.co/datasets/cyberco/caia-0927}{HuggingFace DataSet}}
\vspace{1em}
}

\begin{document}

\maketitle

\thispagestyle{firstpagestyle} 
\renewcommand\thefootnote{}\footnote{$^{*}$ Equal Contributions. $^{\dagger}$ Corresponds to: Ryan Yihe Li, \href{mailto:r@cybertinolab.com}{r@cybertinolab.com}.}

\input{tex/0_abs}

\newpage
\input{tex/1_intro}

\input{tex/2_curation}

\input{tex/3_analysis}

\input{tex/4_conclusion}

\clearpage

\bibliographystyle{plainnat}
\bibliography{references}

\end{document}

%% file: tex/command/math_commands.tex
\usepackage{amsmath,amsfonts,bm}

\def\eqref#1{equation~\ref{#1}}

\def\1{\bm{1}}

\DeclareMathAlphabet{\mathsfit}{\encodingdefault}{\sfdefault}{m}{sl}
\SetMathAlphabet{\mathsfit}{bold}{\encodingdefault}{\sfdefault}{bx}{n}



%% file: tex/command/mlVecMat.tex
\usepackage{amsmath}
\usepackage{amsfonts}
\usepackage{amssymb}
\usepackage{wrapfig}
\usepackage{subcaption}
\usepackage{multirow}
 \usepackage{mathtools} 

\usepackage{verbatim}

\usepackage{anyfontsize}

\usepackage{microtype}
\usepackage{graphicx}
\usepackage{booktabs} 
\usepackage{multirow}
\usepackage{amsmath,amssymb}
\usepackage{booktabs}
\usepackage{caption,subcaption}

\usepackage{xcolor}
\definecolor{mygreen}{HTML}{3cb44b}
\definecolor{skyblue}{HTML}{beffff}
\definecolor{lightgreen}{HTML}{90ee90}

\usepackage{color, colortbl}

\definecolor{emerald}{rgb}{0.31, 0.78, 0.37}

\usepackage{tcolorbox}
\usepackage{enumitem}
\usepackage{colortbl}

\usepackage{xcolor}
\definecolor{mygreen}{HTML}{3cb44b}
\colorlet{myyellow}{green!10!orange!90!}
\makeatletter

\usepackage{tikz}
\usetikzlibrary{arrows,shapes,snakes,automata,backgrounds,fit,petri}
\usepackage{adjustbox}

\newcommand{\RN}[1]{%
	\textup{\lowercase\expandafter{\it \romannumeral#1}}%
}
\usepackage{tabu}

\newcommand{\beq}{\vspace{0mm}\begin{equation}}
\newcommand{\eeq}{\vspace{0mm}\end{equation}}
\newcommand{\beqs}{\vspace{0mm}\begin{eqnarray}}
\newcommand{\eeqs}{\vspace{0mm}\end{eqnarray}}
\newcommand{\barr}{\begin{array}}
\newcommand{\earr}{\end{array}}

\usepackage{color, colortbl}
\definecolor{Gray}{gray}{0.93}

\usepackage{lipsum}

\usepackage{pifont}

\usepackage{makecell}

\usepackage{xcolor,amsmath}
\usepackage[linesnumbered,ruled,vlined]{algorithm2e}
\DontPrintSemicolon

\usepackage{xcolor}
\definecolor{mygreen}{HTML}{3cb44b}

\SetKwComment{Comment}{\color{green!50!black}\# }{}

\SetKwProg{Function}{def}{:}{}

\SetKwProg{For}{for}{:}{}
\SetKwProg{If}{if}{:}{}

%% file: tex/command/box.tex
\definecolor{darkred}{RGB}{140, 21, 21}
\definecolor{citegray}{gray}{0.7}
\definecolor{orange}{HTML}{F58025}

\definecolor{deepred}{rgb}{0.631,0.102,0.102}
\definecolor{amethyst}{rgb}{0.6, 0.4, 0.8}
\definecolor{darkgreen}{rgb}{0.3,0.7,0.3}
\definecolor{salmon}{RGB}{241, 150, 141}
\definecolor{mildyellow}{HTML}{FFF2CC}

\usepackage{xspace}
\tcbuselibrary{breakable}
\usepackage{xcolor}
\usepackage{setspace}

\definecolor{aiblue}{RGB}{66, 133, 244}
\definecolor{humangreen}{RGB}{15, 157, 88}
\definecolor{lightgray}{RGB}{245, 245, 245}
\definecolor{codebg}{RGB}{240, 240, 240}

\newcommand{\airesponsetitle}{}

\newcounter{aimessage}

\newcommand{\humanprompttitle}{}

\newcounter{humanmessage}

%% file: tex/0_abs.tex
\begin{abstract}

We present CAIA, a benchmark exposing \textbf{a critical blind spot in AI evaluation: the inability of state-of-the-art models to operate in adversarial, high-stakes environments where misinformation is weaponized and errors are irreversible}. While existing benchmarks measure task completion in controlled settings, real-world deployment demands resilience against active deception. Using cryptocurrency markets as a natural laboratory, where \$30 billion was lost to exploits in 2024, we evaluate 17 leading models on 178 time-anchored tasks requiring agents to distinguish truth from manipulation, navigate fragmented information landscapes, and make irreversible financial decisions under adversarial pressure.

Our results reveal a fundamental capability gap: without tools, even frontier models achieve only 12-28\% accuracy on tasks junior analysts routinely handle. Tool augmentation improves performance but plateaus at 67.4\% (GPT-5) versus 80\% human baseline, despite unlimited access to professional resources. Most critically,  \textbf{we uncover a systematic tool selection catastrophe: models preferentially choose unreliable web search (55.5\% of invocations) over authoritative blockchain data}, falling for SEO-optimized misinformation and social media manipulation. This behavior persists even when correct answers are directly accessible through specialized tools, suggesting foundational limitations rather than knowledge gaps.

The implications extend beyond cryptocurrency to \textbf{any domain where adversaries actively exploit AI weaknesses}, e.g. cybersecurity, content moderation, etc. Our finding that Pass@k metrics mask dangerous trial-and-error behavior challenges fundamental assumptions about autonomous deployment. We release CAIA with contamination controls and continuous updates, establishing adversarial robustness as a necessary condition for trustworthy AI autonomy. The benchmark reveals that current models, despite impressive reasoning scores, remain fundamentally unprepared for environments where intelligence must survive active opposition.

\end{abstract}

%% file: tex/1_intro.tex
\section{Introduction}

\paragraph{The Gap Between Benchmark Performance and Autonomous Agent Deployment.}

Artificial intelligence benchmarks guide optimization, shape incentives, and define progress in modern AI \cite{imagenet,glue}. Over the past year, foundation models have achieved remarkable milestones: OpenAI models won the International Collegiate Programming Contest \cite{icpc}, and Gemini with DeepThink solved International Mathematical Olympiad problems at gold-medal level \cite{imo}, surpassing most human experts. These achievements have fueled optimism about deploying autonomous AI agents with minimal human oversight. Yet this optimism rests on a dangerous assumption that high scores translate directly to real-world readiness.

Most benchmarks evaluate models in closed worlds where tools function as expected, information is trustworthy, and other agents cooperate \cite{mialon2023gaia,zhou2023webarena,yao2025spin,zheng2025livecodebench}. \textbf{They measure competence, not resilience.} Real-world autonomy requires surviving in open systems rife with uncertainty, misinformation, and adversarial incentives. Agents deployed in finance, governance, or infrastructure must distinguish truth from manipulation, avoid catastrophic failure, and act conservatively under uncertainty. Evaluation of autonomous AI agents, where trustworthy deployment is the top priority, should therefore critical capabilities explicitly.

This gap creates a perilous blind spot in measuring AI progress. An agent that excels on challenging reasoning benchmarks may still believe fabricated news, purchase compromised assets, or fall for phishing attacks, because nothing in its evaluation prepared it for deception. As AI agents increasingly interact with untrusted users, real money, and critical infrastructure, this vulnerability represents a safety concern hiding behind impressive scores \cite{amodei2016concrete,hendrycks2021unsolved}.

We argue for an opinion shift in agent evaluation: \textbf{Beyond measuring task completion on curated problems, evaluations should test robust survival in adversarial, high-stakes environments}. Rather than escalating difficulty alone, we should simulate hostile settings where others actively deceive, information is weaponized, and irreversible failures cause substantial loss. We introduce \emph{CAIA}, the Crypto AI Agent Benchmark, which tests AI agent capabilities under these conditions.

\paragraph{Crypto: A Natural Laboratory for Adversarial Robustness.}

Cryptocurrency markets provide a unique environment for evaluating agent robustness under genuinely adversarial conditions. Despite controversy around speculation and fraud, these characteristics create ideal hostile testing conditions for AI agents. Crypto uniquely combines three properties essential for adversarial evaluation:

\textbf{1. Adversarial Environment with Sophisticated Deception.} The cryptocurrency ecosystem operates as a ``dark forest" where misinformation is weaponized and adversaries actively hunt victims \cite{darkforest}. Pseudonymous blockchains enable malicious actors to operate without reputation consequences. Potential profits motivate sophisticated attack strategies. Regulatory gaps permit deception tactics illegal in traditional markets. Daily occurrences include honeypot contracts designed to trap victims \cite{honeypot_analysis}, flash loan exploits manipulating prices within single transactions \cite{flashloan_attacks}, and coordinated social engineering campaigns \cite{scam_detection}. These real adversarial conditions require agents to genuinely distinguish truth from manipulation.

\textbf{2. High Stakes with Immediate Consequences.} Cryptocurrency markets lack traditional financial safeguards. Transactions are irreversible, smart contract executions are final, and no central authority can reverse fraudulent transfers. In 2024 alone, over \$30 billion was lost to exploits and scams \cite{chainalysis2025}. When an AI agent makes a tiny mistake, losses cannot be recovered. This creates genuine high-stakes conditions where errors have immediate, permanent monetary consequences and malicious actors are economically incentivized to exploit weaknesses.

\textbf{3. Transparent and Verifiable Ground Truth.} Despite adversarial chaos, cryptocurrency offers complete transparency and immutability. Every transaction, smart contract interaction, and token transfer is permanently recorded on public blockchains \cite{blockchain_data_challenges}. This enables unique evaluation conditions where: (1) agent decisions can be verified against immutable on-chain records; (2) financial losses trace to specific transactions with cryptographic proof; (3) attack patterns can be analyzed retroactively with perfect information. This transparency in an adversarial environment enables reproducible evaluation with real-world relevance, addressing fundamental limitations of traditional financial benchmarks that must choose between proprietary data or synthetic simulations.

Current AI systems enter this domain fundamentally unprepared. Trained predominantly on centralized, indexed, trustworthy ``Web2" data \cite{commoncrawl,c4}, they lack exposure to crypto's fragmented, rapidly-evolving ``Web3" information landscape. Blockchain data spans thousands of nodes without central access points; DeFi protocols update daily without documentation; critical information exists in ephemeral social channels that evade crawlers \cite{blockchain_data_challenges}. Even accessible content is often adversarial, consisting of deliberately misleading information, scams, and market manipulation. This combination makes crypto particularly challenging for AI trained on traditional web data, and hence an ideal testbed for AI agents' adversarial robustness.

\paragraph{CAIA: Benchmarking Intelligence Under Fire.}

We present CAIA (Crypto AI Agent Benchmark), the first benchmark explicitly designed to evaluate AI agents in an actively hostile, high-stakes environment. Unlike existing benchmarks measuring task completion in controlled settings, all tasks in CAIA are grounded in crypto, which measure survival and truth-seeking under adversarial pressure.

Our evaluation reveals significant gaps between state-of-the-art large language models and junior human analysts. Models achieve only 12-28\% accuracy without tools. Even when equipped with tools providing correct answers,  the accuracy at best is 67.4\% (\texttt{GPT-5}), while entry-level human analyst baselines reach 80\%. Models consistently rely on unreliable web search over domain-specific tools that directly link to the source of truth, suggesting fundamental limitations in tool selection and adversarial reasoning. These patterns reveal that, when users entrust capital to autonomous agents expecting intelligent fund management, agents may effectively be guessing and attempting ``trial-and-error", which is extremely dangerous in high-stakes adversarial scenarios.

\paragraph{Our Contributions.}

Our work advances agent evaluation through four primary contributions:

\textbf{Adversarial-First Evaluation:} While existing benchmarks assume cooperative environments, CAIA introduces active deception, source validation, and adversarial robustness as core capabilities, reflecting deployment reality where agents face hostile actors, not just noisy data.

\textbf{Financial Reality Grounding:} Using real market tasks where mistakes have monetary consequences creates accountability and real-world transferability absent from synthetic benchmarks.

\textbf{Temporal Precision Testing:} Time-anchored tasks evaluate multi-timescale reasoning and data obsolescence handling required in volatile markets, beyond static benchmark capabilities.

\textbf{Diagnostic Failure Analysis:} Fine-grained diagnostics of evaluation results provide actionable insights about specific failure modes, critical for both model development and deployment decisions.

The implications extend beyond crypto. As AI agents enter other adversarial domains, e.g. cybersecurity, content moderation, medical diagnosis, CAIA's measured capabilities become universally critical. Crypto represents an extreme adversarial environment characterized by pervasive misinformation, sophisticated scams, and active financial exploits. Success on CAIA therefore provides high confidence for autonomous deployment in any domain where adversaries actively exploit weaknesses, and establishes a strong foundation for routine deployment in less hostile environments.

\paragraph{Paper Organization.}
In the following, Section \ref{curation} presents CAIA's design philosophy and task curation methodology. Section \ref{evaluation} details our experimental framework and quantitative evaluation results across 17 state-of-the-art models. Section \ref{analysis} analyzes failure modes and derives insights for improving and deploying AI agents. Section \ref{conclusion} discusses future directions and concludes the paper.

%% file: tex/2_curation.tex
\section{Benchmark Curation}\label{curation}

\subsection{Design Principles}

CAIA addresses a critical gap in agent evaluation: the absence of benchmarks that capture the worst-case performance under adversarial, high-stakes scenarios, which is exactly the nature of real-world crypto analysis. We identify three core challenges that define this domain: 
\begin{itemize}
    \item irreversible financial consequences where incorrect decisions lead to permanent capital loss (e.g., MEV and execution risks)~\cite{daian2019flashboys,qin2021bev};
    \item an adversarial information landscape, including coordinated manipulation and pump-and-dump campaigns~\cite{xu2019anatomy,ardia2023twitter}; 
    \item high-density, multi-source data that mixes on-chain traces, social signals, and protocol documentation~\cite{mialon2023gaia,zhou2023webarena}.
\end{itemize}

Our community-driven curation process, involving over 3,000 contributors including protocol developers, quantitative researchers, and venture capital investors, ensures ecological validity.

To mitigate training-data contamination, a persistent threat to static benchmarks~\cite{deng2024contamination,carlini2021extracting}, CAIA anchors tasks to recent market events with explicit temporal constraints (block heights, timestamps), following best practices from time-sensitive evaluation~\cite{chen2021timeqa,kasai2023realtimeqa}, and creating an evaluation framework resistant to memorization-based solutions. We will also actively retire out-dated tasks and add new tasks to ensure liveness.

\subsection{Quality Assurance}\label{quality}

For each task in CAIA, quality is guaranteed through three foundational pillars that mirror expert analytical workflows. This approach moves beyond isolated capability testing to evaluate complete reasoning and acting chains~\cite{yao2022react}, ensuring that successful task completion requires:

\textbf{Knowledge:} Evaluates foundational understanding of crypto-native concepts, from AMM mechanics to governance structures, testing conceptual grasp rather than definitional recall.

\textbf{Planning:} Assesses strategic decomposition of complex questions into executable analytical workflows, requiring agents to specify tool selection and sequencing before execution~\cite{wei2022chainofthought}.

\textbf{Action:} Tests real-world execution using production APIs (Etherscan, CoinGecko, DefiLlama)~\cite{etherscan_docs,coingecko_docs,defillama_docs} dedicated for on-chain detection, evaluating both technical competence and judgment under realistic constraints like rate limiting and data inconsistency.

This progression from understanding through planning to execution reflects established cognitive architectures in complex problem-solving~\cite{newell1990unified} specifically adapted for the crypto domain's unique requirements. Tasks that satisfy these conditions are naturally suited for testing AI agents' capabilities, as they precisely mirror the desired approach to tackling complex problems.

\subsection{Curation Pipeline}

Our dataset originates from more than 10,000 authentic queries we collect from over 3,000 active users spanning different roles, representing the largest and most comprehensive collection of real-world crypto analytical needs to date. Through a rigorous five-stage curation pipeline that operationalizes our design principles and quality metrics (Figure~\ref{fig:curation}), we distill the candidate pool into 178 high-quality CAIA benchmark tasks, as detailed below:

\begin{figure*}[htbp]
    \centering
    \includegraphics[width=\linewidth]{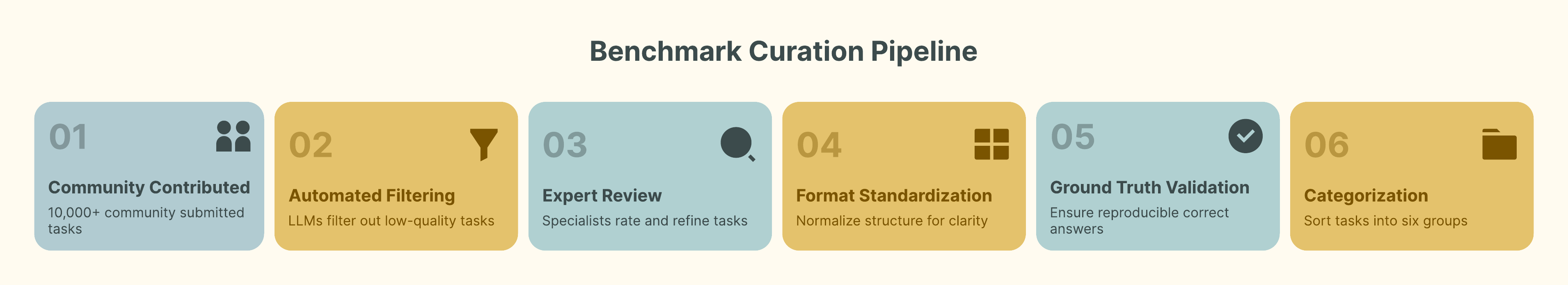}
    \caption{Data Curation Pipeline.}
    \label{fig:curation}
\end{figure*}

\textbf{Stage 1: Automated Filtering:} We apply the standard ``LLM-as-a-judge" technique, using LLMs to filter out off-topic, ambiguous, non-answerable, or trivial queries while enforcing temporal grounding. After filtration, we ask LLMs to rate each task based on our quality assurance criteria in \ref{quality}, retaining only the top 15\%. This reduces the corpus size to approximately 1,000 tasks.

\textbf{Stage 2: Expert Review:} This stage mirrors the traditional paper reviewing process. Our expert team comprises 92 domain specialists, with each assigned 50 tasks to review and grade based on the quality assurance criteria in \ref{quality}. Each task surviving Stage 1 receives at least 4 reviews, and we calculate the final score by averaging all reviews after removing the highest and lowest scores. The top 200 tasks advance to the final pool. After deduplicating similar tasks (i.e., handpicking tasks requiring similar execution logic), we obtain a prototype candidate set of 186 tasks.

\textbf{Stage 3:  Format Standardization:} To address inconsistencies arising from different tones and writing styles, we unify the format of each task. This requires explicit anchoring to block numbers or timestamps, enabling objective evaluation and straightforward verification of ground truth answers.

\textbf{Stage 4: Ground Truth Validation:} For each task and its corresponding answer, we verify that a reproducible ground truth toolchain calling scheme exists and associate it with the task. We omit tasks that cannot be reproduced from the benchmark to ensure objectivity. This process provides much more than a single correct answer - It demonstrates the precise methodology that agents should follow to reach the correct solution, ensuring the accuracy, objectivity, and reproducibility of the desired answer. After this validation, we arrive at our final CAIA benchmark of 178 tasks.

\textbf{Stage 5: Categorization:} For diagnostics of model capabilities, we identify 6 fine-grained categories encompassing all tasks and carefully categorize each task accordingly, as shown in Table \ref{tab:task_distribution}. This step enables detailed assessment beyond aggregate metrics, supporting our analysis in Section \ref{analysis}.

\begin{table*}[h]
\centering
\small
\begin{tabular}{lccll}
\toprule
\textbf{Category} & \textbf{N} & \textbf{\%} & \textbf{Focus} & \textbf{Validation Method} \\
\midrule
On-Chain Analysis & 77 & 43.3 & Transaction patterns, MEV, fund flows & Transaction hash verification \\
Project Discovery & 49 & 27.5 & Protocol evaluation, security analysis & Documentation cross-reference \\
Tokenomics & 23 & 12.9 & Incentive design, value accrual & Mathematical proof \\
Overlap & 14 & 7.9 & Multi-domain synthesis & Composite verification \\
Trend Analysis & 8 & 4.5 & Temporal patterns, adoption metrics & Statistical validation \\
General Knowledge & 7 & 3.9 & Foundational concepts & Canonical reference \\
\bottomrule
\end{tabular}
\caption{Distribution of 178 benchmark tasks across 6 analytical categories.}
\label{tab:task_distribution}
\end{table*}

By design, CAIA addresses weaknesses noted in prior evaluations: contamination in static datasets~\cite{deng2024contamination}, lack of ecological validity in synthetic tasks~\cite{zhou2023webarena,mialon2023gaia}, and single-metric reporting that masks capability gaps~\cite{liang2022helm}. Grounding in real-world high-stakes needs, operating under an adversarial environment with false information, verifiable with objective and immutable answers, CAIA provides a durable foundation for measuring autonomous agentic intelligence in adversarial financial markets.

%% file: tex/3_analysis.tex
\section{Evaluation Results}\label{evaluation}

\subsection{Experimental Setup}

We conduct a comprehensive evaluation of 17 state-of-the-art large language models on the CAIA benchmark, encompassing leading proprietary models (\texttt{GPT-4.1}, \texttt{GPT-5}, \texttt{Claude}, \texttt{Gemini}, \texttt{Grok}, \texttt{Kimi}) and prominent open-source flagships (\texttt{Llama}, \texttt{Qwen}, \texttt{DeepSeek}, \texttt{GPT-OSS}). 

\textbf{Tool Augmentation.} We evaluate each model under two distinct conditions that mirror complementary aspects of real-world deployment. The \textbf{without-tools} condition functions as a closed-book examination, testing models' internalized knowledge and reasoning capabilities when forced to rely solely on parametric memory. This reveals their fundamental understanding of concepts, market dynamics, and analytical reasoning without external assistance. Conversely, the \textbf{with-tools} condition resembles an open-book examination and tests agentic abilities where models gain access to 23 specialized tools spanning web search APIs, blockchain analytics platforms, market data feeds, and computational interpreters. Crucially, our data curation process in \ref{curation} ensures that correct answers are always accessible through appropriate tool use, and thus the challenge lies not in information availability but in tool selection and synthesis. This design choice deliberately isolates the agent's tool orchestration capabilities from knowledge limitations, providing a pure test of whether agents can identify and invoke the right resources when given unlimited access to professional instruments.

\textbf{Agentic Framework.} When equipped with tools, each model operates within a standardized ReAct-style \cite{yao2022react} agentic framework that handles tool dispatch, result parsing, and iterative reasoning. This ensures that our evaluation result is not affected by implementation variations.

\textbf{Human Baseline.} To establish human performance benchmarks, we recruited 16 participants from university blockchain clubs and early-stage blockchain companies, representing entry-level analyst expertise. These participants completed a stratified 10\% sample of our benchmark, carefully balanced across all six analytical domains. Their averaged performance of 80\% accuracy provides a critical baseline---notably, these junior analysts achieved this score in the open-book equivalent condition with full tool access, establishing the minimum bar for professional competence in the domain.
\subsection{Quantitative Performance Analysis}

To ensure robust evaluation, we employ multiple complementary metrics. Our primary measure is average accuracy via majority voting across five independent runs, which mitigates the substantial variance inherent in single-run evaluations of stochastic language models. We additionally report standard Pass@1 and Pass@5 metrics to capture both first-attempt performance and eventual success rates through exploration. However, as we will discuss in Section~\ref{analysis}, the traditional Pass@k is misleading in high-stakes adversarial contexts where trial-and-error carries unacceptable risks.

Beyond performance metrics, we track computational costs by logging token consumption for each query and computing the associated monetary expense. This enables us to derive cost efficiency (cost-per-accuracy-point), revealing critical trade-offs between model capability and economic viability. This analysis proves particularly illuminating when comparing proprietary APIs against open-source alternatives, where we observe up to 100-fold differences in cost for comparable performance.

\begin{figure*}[htbp]
    \centering
    \includegraphics[width=.8\linewidth]{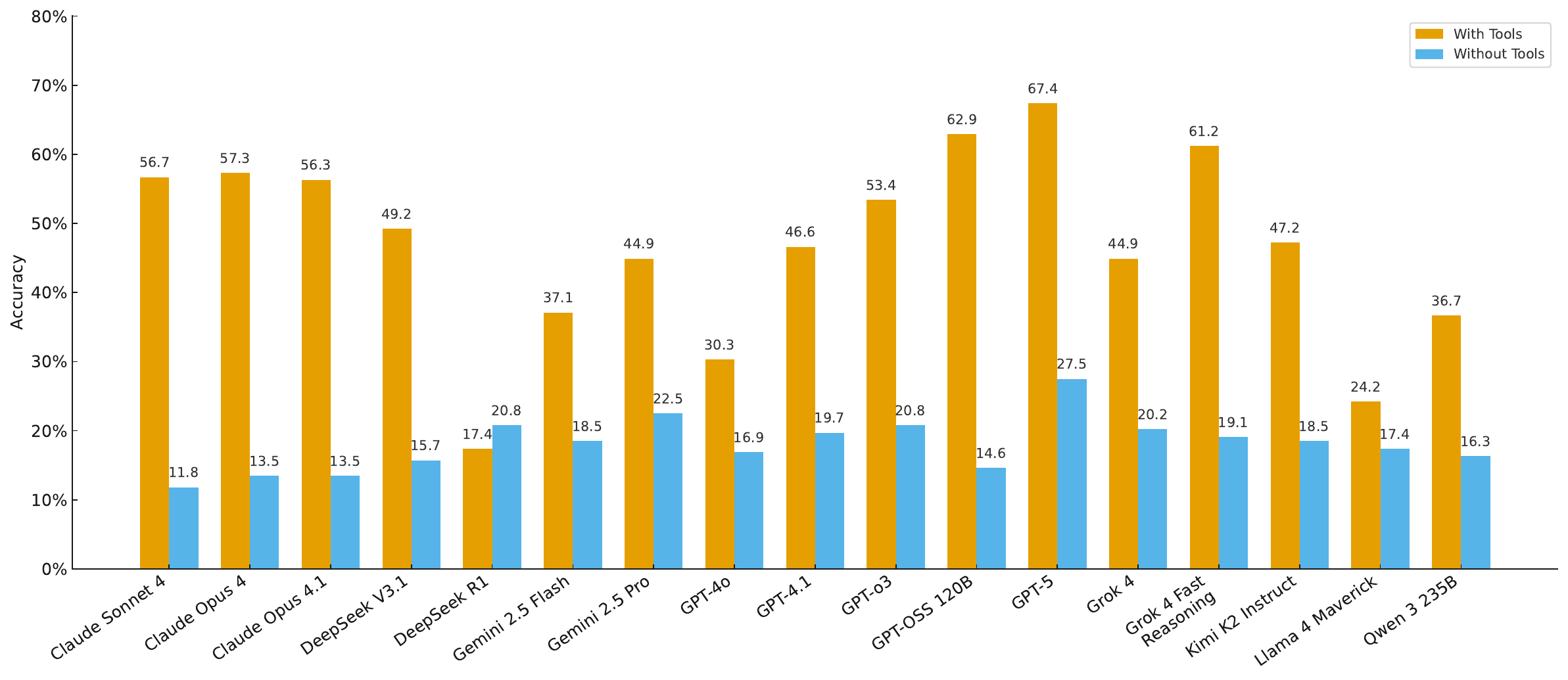}
    \caption{Average accuracy across five evaluation runs using majority voting. The dashed line indicates 80\% human baseline performance. Without tools, all models perform near random (12--28\%); with tools, performance improves but plateaus below human capability.}
    \label{fig:model_scores}
\end{figure*}

Our results reveal a stark performance landscape that challenges fundamental assumptions about tool-augmented language models. As illustrated in Figure~\ref{fig:model_scores} and detailed in Tables~\ref{tab:without-tools-merged} and~\ref{tab:with-tools-merged}, model performance exhibits a bimodal distribution heavily dependent on tool availability.

In the absence of external tools, every evaluated model, including frontier systems, demonstrates catastrophic failure, achieving merely 12--28\% accuracy. This represents performance scarcely above random guessing for many tasks. Even \texttt{GPT-5}, the strongest model, has only 27.5\% accuracy, indicating how poorly current parametric knowledge transfers to specialized adversarial domains.

On the other hand, tool augmentation yields substantial improvements, yet even our best-performing model \texttt{GPT-5} achieves only 67.4\% accuracy, falling significantly short of the 80\% human baseline established by junior analysts. This performance ceiling persists despite unlimited access to professional-grade tools and comprehensive documentation, suggesting fundamental architectural limitations of the state-of-the-art LLMs today, rather than simple knowledge gaps.

\begin{figure*}[htbp]
    \centering
    \includegraphics[width=0.7\linewidth]{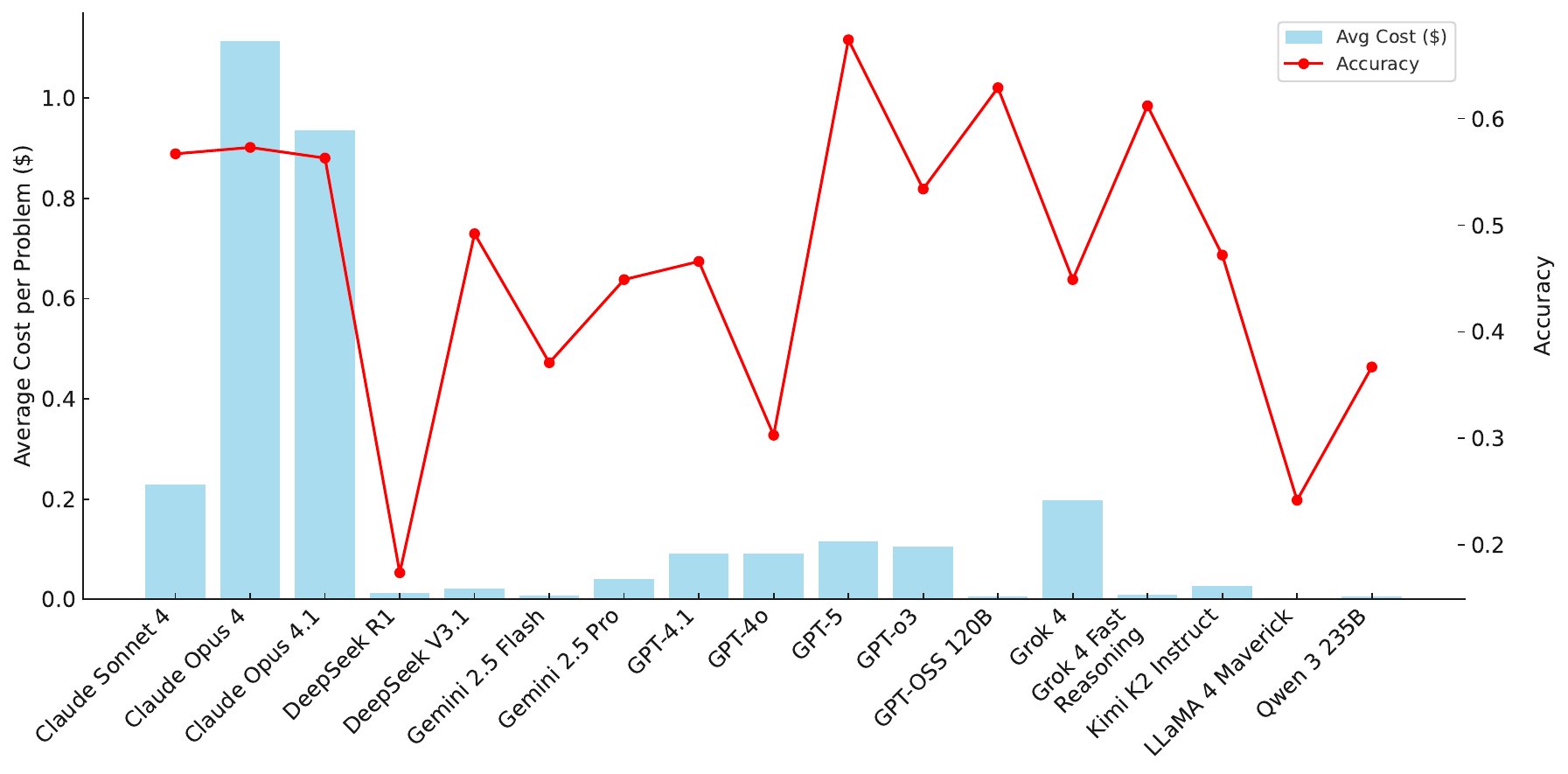}
    \caption{Cost-accuracy tradeoff reveals GPT-OSS 120B and Grok 4 Fast as Pareto-optimal choices, achieving near-frontier performance at minimal cost.}
    \label{fig:cost_accuracy_with_tool}
\end{figure*}

The cost-efficiency analysis reveals a striking economic disparity across model families. While proprietary systems like \texttt{Claude Opus 4} incur costs exceeding \$1 per problem, open-source alternatives such as \texttt{GPT-OSS 120B} achieve competitive accuracy at under \$0.01 per query, which is a remarkable 100-fold improvement in cost efficiency. Even more compelling, \texttt{GPT-OSS 120B} actually \emph{outperforms} several proprietary models while maintaining this dramatic cost advantage. This economic reality has profound implications for deployment at scale: organizations processing thousands of queries daily could achieve near-frontier performance at a fraction of the cost, fundamentally challenging the assumed superiority of commercial APIs in specialized domains, as illustrated in  \ref{fig:cost_accuracy_with_tool}.

\begin{table*}[htbp]
\centering
\small
\setlength{\tabcolsep}{8pt}
\renewcommand{\arraystretch}{1.15}
\begin{tabular}{
  l
  S[table-format=1.3]
  S[table-format=2.1]
  S[table-format=2.1]
  S[table-format=1.4]
  S[table-format=1.4]
}
\toprule
\textbf{Model} & \textbf{Majority Vote} & \textbf{Pass@1 (\%)} & \textbf{Pass@5 (\%)} & \textbf{Avg Cost (\$)} & \textbf{Cost/Score} \\
\midrule
Claude Sonnet 4        & 0.118 & 12.9 & 18.0 & 0.0070 & 0.0593 \\
Claude Opus 4          & 0.135 & 13.5 & 16.9 & 0.0334 & 0.2481 \\
Claude Opus 4.1        & 0.135 & 15.2 & 17.4 & 0.0356 & 0.2642 \\
\midrule
DeepSeek R1            & 0.208 & 21.9 & 35.4 & 0.0038 & 0.0184 \\
DeepSeek V3.1          & 0.157 & 15.7 & 29.2 & 0.0005 & 0.0030 \\
\midrule
Gemini 2.5 Flash       & 0.185 & 20.2 & 21.9 & 0.0012 & 0.0062 \\
Gemini 2.5 Pro         & 0.225 & 20.2 & 29.8 & 0.0051 & 0.0226 \\
\midrule
GPT-4.1                & 0.197 & 20.8 & 24.2 & 0.0025 & 0.0126 \\
GPT-4o                 & 0.169 & 19.1 & 20.8 & 0.0016 & 0.0098 \\
GPT-5                  & 0.275 & 28.1 & 42.7 & 0.0207 & 0.0753 \\
GPT-o3                 & 0.208 & 22.5 & 29.2 & 0.0085 & 0.0407 \\
GPT-OSS 120B           & 0.146 & 18.5 & 21.3 & 0.0003 & 0.0022 \\
\midrule
Grok 4                 & 0.202 & 20.2 & 24.2 & 0.0345 & 0.1705 \\
Grok 4 Fast Reasoning  & 0.191 & 21.3 & 23.6 & 0.0006 & 0.0029 \\
\midrule
Kimi K2 Instruct       & 0.185 & 17.4 & 25.3 & 0.0006 & 0.0033 \\
\midrule
Llama 4 Maverick       & 0.174 & 16.9 & 24.7 & 0.0003 & 0.0015 \\
\midrule
Qwen 3 235B            & 0.163 & 14.6 & 18.0 & 0.0010 & 0.0061 \\
\bottomrule
\end{tabular}
\caption{Performance \emph{without tools}: accuracy, Pass@k, cost, and cost efficiency across all models.}
\label{tab:without-tools-merged}
\end{table*}

\begin{table*}[htbp]
\centering
\small
\setlength{\tabcolsep}{8pt}
\renewcommand{\arraystretch}{1.15}
\begin{tabular}{
  l
  S[table-format=1.3]
  S[table-format=2.1]
  S[table-format=2.1]
  S[table-format=1.4]
  S[table-format=1.4]
}
\toprule
\textbf{Model} & \textbf{Majority Vote} & \textbf{Pass@1 (\%)} & \textbf{Pass@5 (\%)} & \textbf{Avg Cost (\$)} & \textbf{Cost/Score} \\
\midrule
Claude Sonnet 4        & 0.567 & 57.9 & 66.9 & 0.2291 & 0.4037 \\
Claude Opus 4          & 0.573 & 59.6 & 71.9 & 1.1139 & 1.9439 \\
Claude Opus 4.1        & 0.563 & 56.3 & 69.0 & 0.9357 & 1.6614 \\
\midrule
DeepSeek R1            & 0.174 & 26.4 & 54.5 & 0.0121 & 0.0695 \\
DeepSeek V3.1          & 0.492 & 55.9 & 71.2 & 0.0216 & 0.0438 \\
\midrule
Gemini 2.5 Flash       & 0.371 & 39.3 & 62.4 & 0.0070 & 0.0190 \\
Gemini 2.5 Pro         & 0.449 & 49.4 & 61.2 & 0.0407 & 0.0906 \\
\midrule
GPT-4.1                & 0.466 & 51.7 & 60.7 & 0.0913 & 0.1958 \\
GPT-4o                 & 0.303 & 50.0 & 55.6 & 0.0909 & 0.2997 \\
GPT-5                  & 0.674 & 70.2 & 77.0 & 0.1154 & 0.1712 \\
GPT-o3                 & 0.534 & 59.6 & 73.6 & 0.1047 & 0.1962 \\
GPT-OSS 120B           & 0.629 & 56.2 & 72.5 & 0.0066 & 0.0104 \\
\midrule
Grok 4                 & 0.449 & 52.2 & 66.9 & 0.1980 & 0.4405 \\
Grok 4 Fast Reasoning  & 0.612 & 57.9 & 71.9 & 0.0098 & 0.0160 \\
\midrule
Kimi K2 Instruct       & 0.472 & 46.6 & 64.6 & 0.0273 & 0.0579 \\
\midrule
Llama 4 Maverick       & 0.242 & 30.3 & 64.6 & 0.0031 & 0.0129 \\
\midrule
Qwen 3 235B            & 0.367 & 38.4 & 61.0 & 0.0062 & 0.0170 \\
\bottomrule
\end{tabular}
\caption{Performance \emph{with tools}: accuracy, Pass@k, cost, and cost efficiency across all models.}
\label{tab:with-tools-merged}
\end{table*}

\section{Analysis and Discussion}\label{analysis}

\subsection{The Illusion of Competence: Why Pass@k Metrics Mislead}

A critical finding emerges from the stark divergence between Pass@1 and Pass@5 metrics shown in Table~\ref{tab:with-tools-merged}. While Pass@1 accuracy hovers around 50--60\% for top models except for 70.2\% of \texttt{GPT-5}, Pass@5 consistently exceeds 60\%, with \texttt{GPT-5} reaching 77\%. This improvement might initially suggest robust problem-solving capabilities. However, this interpretation fundamentally misunderstands the nature of real-world deployment, particularly in high-stakes  financial contexts.

Consider the implications: \texttt{Gemini 2.5 Flash}'s jump from 39.3\% (Pass@1) to 62.4\% (Pass@5) indicates that the model essentially \emph{guesses} correctly through repeated trials rather than reasoning strategically on first attempt. In cryptocurrency markets, where a single incorrect transaction can result in irreversible financial loss, this trial-and-error approach represents an unacceptable risk profile. \textbf{Real-world financial decisions do not offer multiple attempts.} When users entrust capital to autonomous agents, expecting them to manage funds ``cleverly'' through clear reasoning, it is unacceptable and extremely dangerous if the agent is effectively guessing its next action.

The minimal improvement (and sometimes decrease) from Pass@1 to majority voting further underscores this concern. If models were exhibiting genuine understanding with occasional errors, we would expect majority voting to substantially improve accuracy. Instead, the modest gains suggest that errors stem from fundamental reasoning failures rather than stochastic variations. This pattern is particularly alarming given that automated agents are increasingly deployed in financial contexts where users may place unwarranted trust in their recommendations without human oversight.

\subsection{Domain-Specific Tool Usage and Performance Patterns}

\begin{figure*}[htbp]
    \centering
    \begin{minipage}[t]{0.48\linewidth}
        \centering
        \includegraphics[width=.80\linewidth]{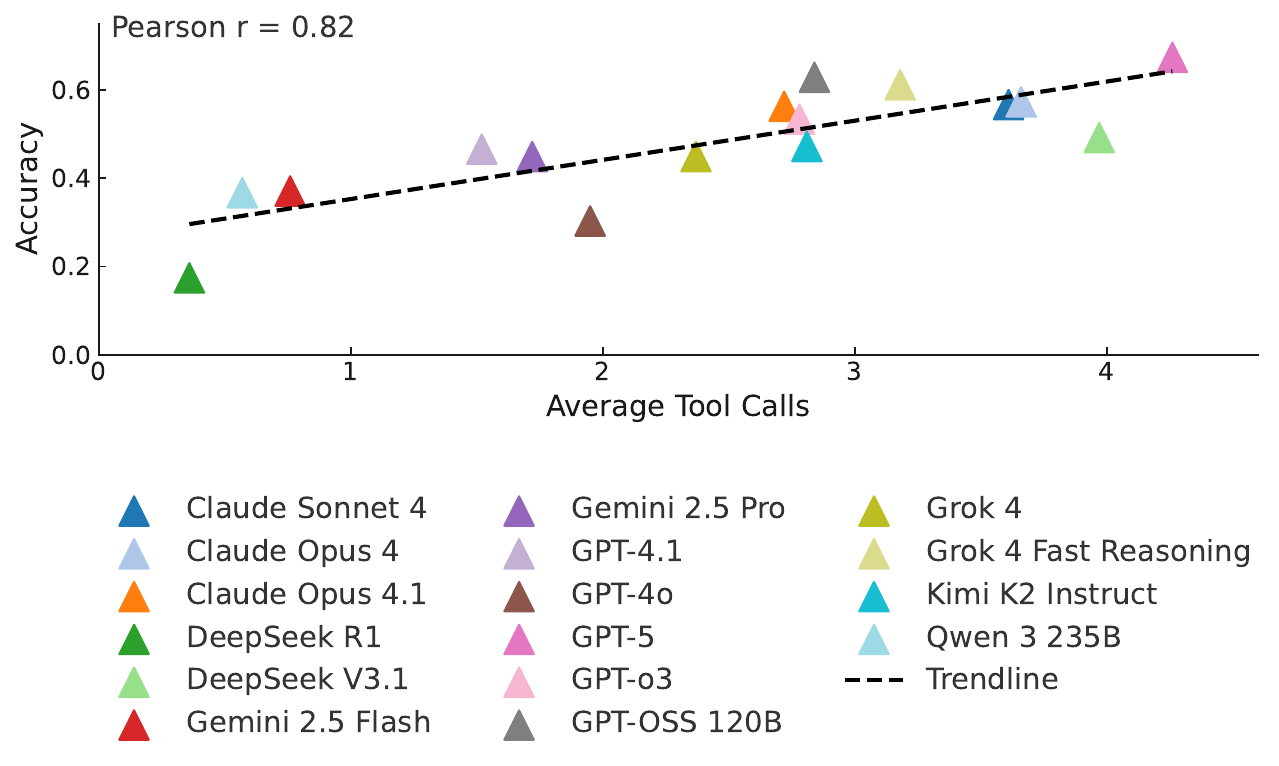}
        \caption{Tool usage frequency vs. accuracy.}
        \label{fig:avg_tools_bar}
    \end{minipage}%
    \hfill
    \begin{minipage}[t]{0.48\linewidth}
        \centering
        \includegraphics[width=.80\linewidth]{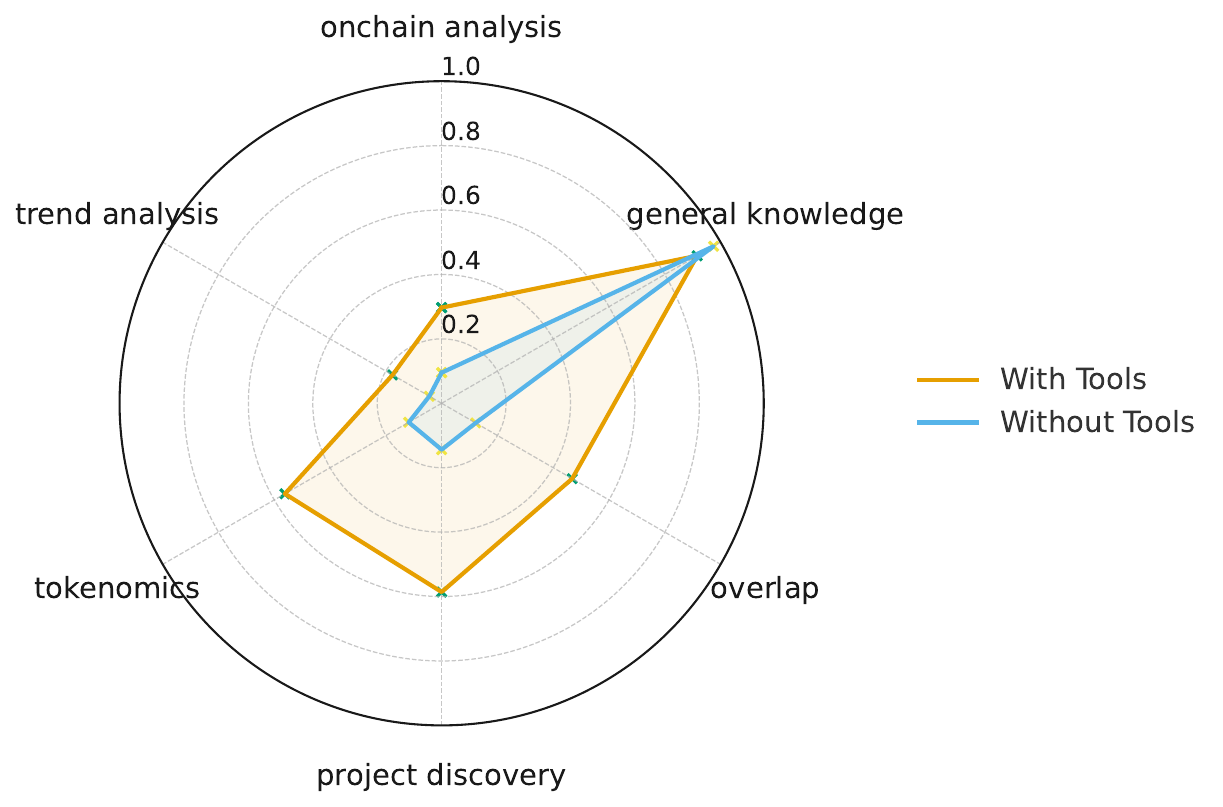}
        \caption{Performance on different categories.}
        \label{fig:category_score}
    \end{minipage}
\end{figure*}

Figure \ref{fig:avg_tools_bar} reveals a positive correlation between tool usage and accuracy with a Pearson coefficient of $0.82$, showing that statistically more tool calls can help iterate and refine the response. On the other hand, the improvement by more tool calls is not apparent, which suggests that effective tool use depends on strategic selection rather than quantity. Models making numerous unfocused tool calls may perform worse than those making fewer, well-targeted queries to appropriate tools.

Figure \ref{fig:category_score} illustrates how this tool effectiveness varies dramatically across CAIA's six analytical categories. In \textbf{general knowledge} tasks, tools provide minimal benefit since this stable, universal knowledge is already well-represented in pre-training corpora, and models perform consistently well with or without tool access. Conversely, \textbf{on-chain analysis} and \textbf{trend analysis} show the largest performance gaps between tool-assisted and non-tool scenarios, yet remain the lowest-performing categories overall. These domains demand not just tool access but sophisticated reasoning about which tools to deploy and how to interpret their outputs within dynamic  environments.

The intermediate categories, \textbf{tokenomics}, \textbf{project discovery}, and \textbf{overlap}, demonstrate the most successful tool integration, with substantial performance improvements when external resources are available. These domains benefit from tools because they are underrepresented in pre-training data, allowing targeted information retrieval to compensate for knowledge gaps.

However, as we reveal in the following subsection, these improvements are largely driven by generic web search rather than specialized tools. LLMs essentially get lucky on these topics because malicious actors have less economic incentive to manipulate information in these relatively obscure domains.

\subsection{The Tool Selection Catastrophe}

From our evaluation, we identify the systematic failure of models to select appropriate tools, even when optimal choices are unambiguous. Table~\ref{tab:tool_usage_grouped}  reveals that models default to generic web search for 55.5\% of all tool invocations (combining Google and Twitter searches), despite having access to specialized blockchain analytics tools that provide authoritative data and direct answer.

\begin{table*}[htbp]
\centering
\begin{tabular}{lrr}
\hline
\textbf{Tool Category} & \textbf{Invocations} & \textbf{Percentage} \\
\hline
Google search & 11,626 & 49.6\% \\
Specialized blockchain tools & 8,351 & 35.6\% \\
URL fetching & 1,743 & 7.4\% \\
Twitter search & 1,388 & 5.9\% \\
Code execution & 355 & 1.5\% \\
\hline
\textbf{Total} & 23,463 & 100.0\% \\
\end{tabular}
\caption{Tool usage distribution reveals heavy reliance on generic search over domain-specific tools.}
\label{tab:tool_usage_grouped}
\end{table*}

This behavior pattern is not merely suboptimal: it is dangerous. In an adversarial and manipulated environment, web search returns manipulated social media posts, coordinated shilling campaigns, and deliberately false information. Meanwhile, blockchain data provides immutable, verifiable ground truth. Yet models consistently choose the unreliable source over the authoritative one.

Our analysis reveals that certain tools require orchestration to be effective. Twitter search accuracy plummets from 40.7\% when used in combination to 6.6\% when used alone, indicating that social sentiment tools need market context to provide value. Conversely, direct blockchain queries (e.g., ERC-20 token info) maintain high accuracy in isolation. Models fail to recognize these compositional requirements, treating all tools as functionally equivalent, and problematically have a preference towards generic search tools, which may deliver second-hand manipulated information, over domain-specific tools that directly provide the source of truth for each query. 

\paragraph{A Case Study: When Simple Tasks Become Impossible.}

Task 49 in CAIA epitomizes the depth of model failure in tool selection. The task requires retrieving monthly token launch counts from Pump.fun. The data is readily available through a single blockchain analytics API call, and the ground truth solution is trivial:

\textsc{defillama\_pump\_stats(month=``2025-01", metric=``launches")}

Yet across all 17 evaluated models, \textbf{not a single one succeeded}. Instead, we observed a consistent pattern of cascading failure where models fall for misinformation: 

\begin{enumerate}
    \item Initial web searches return SEO-optimized but outdated blog posts
    \item Refined searches for specific months yield social media speculation rather than data
    \item Desperation leads to Twitter searches, surfacing coordinated misinformation
    \item  Models synthesize incorrect answers from these unreliable sources
\end{enumerate}

The models never attempt to use DeFiLlama, Dune Analytics, or any blockchain-specific tool, despite these being explicitly documented and available. This represents not just a failed execution but a fundamental inability to recognize when specialized tools are necessary and identify source of truth.

%% file: tex/4_conclusion.tex
\section{Conclusion}\label{conclusion}

We introduce CAIA, the first benchmark evaluating AI agents in high-stakes, adversarial environments. Our evaluation of 17 state-of-the-art models reveals critical gaps: leading models achieve only 67.4\% accuracy with tools versus 80\% human baseline, consistently preferring unreliable web search over specialized blockchain tools. The key obstacle is not tool access but fundamental lack of skeptical reasoning. Agents are easily misled by manipulation and confidently hallucinate critical data.

These vulnerabilities extend beyond crypto to any adversarial domain where misinformation is weaponized. Current models remain dangerously unreliable when stakes are high and adversaries are present. For trustworthy autonomy, future work should prioritize adversarial robustness over task completion metrics, and CAIA provides a vital testbed for building truly reliable autonomous agents.